\pdfoutput=1

\documentclass[11pt]{article}

\usepackage{acl}

\usepackage{times}
\usepackage{latexsym}

\usepackage[T1]{fontenc}

\usepackage[utf8]{inputenc}

\usepackage{microtype}

\usepackage{inconsolata}

\usepackage{listings}
\usepackage{upgreek}
\usepackage{booktabs}
\usepackage[utf8]{inputenc}
\usepackage{tabularx}
\usepackage{booktabs}
\usepackage{makecell}
\usepackage{hyperref}
\usepackage{tikz}

\title{Exploring Prompting Large Language Models as Explainable Metrics}

\author{Ghazaleh Mahmoudi  \\
School of Computer Engineering\\
Iran University of Science and Technology, Iran\\
\href{mailto:gh_mahmoodi@comp.iust.ac.ir}{gh\_mahmoodi@comp.iust.ac.ir}}

\begin{document}
\maketitle
\begin{abstract}
This paper describes the IUST NLP Lab submission to the Prompting Large Language Models as Explainable Metrics Shared Task at the Eval4NLP 2023  Workshop on Evaluation \& Comparison of NLP Systems. We have proposed a zero-shot prompt-based strategy for explainable evaluation of the summarization task using Large Language Models (LLMs). The conducted experiments demonstrate the promising potential of LLMs as evaluation metrics in Natural Language Processing (NLP), particularly in the field of summarization. Both few-shot and zero-shot approaches are employed in these experiments. The performance of our best provided prompts achieved a Kendall correlation of 0.477 with human evaluations in the text summarization task on the test data. Code and results are publicly available on \href{https://github.com/ghazaleh-mahmoodi/Exploring_Prompting_LLMs_AS_Explainable_Metrics}{GitHub}
\footnote{\href{https://github.com/ghazaleh-mahmoodi/Exploring_Prompting_LLMs_AS_Explainable_Metrics}{https://github.com/ghazaleh-mahmoodi/Prompting\_LLMs\_AS\_Explainable\_Metrics}}.

\end{abstract}

\section{Introduction}

 Summarization is crucial for quickly understanding large textual documents. The goal of text summarization is to condense lengthy documents into a concise, coherent, and easily understandable format while retaining the essential information from the original source. However, assessing the quality and performance of summarization systems has proven to be a challenging task.
Commonly used evaluation methods for summarization, such as ROUGE scores \cite{lin-2004-rouge}, have certain limitations. They fail to capture the overall quality, coherence, and interpretability of summaries. Additionally, they rely on human-generated reference summaries, which are time-consuming and subjective. Given the limitations of traditional evaluation approaches, it is important to explore alternative evaluation methods that leverage the capabilities of LLMs and offer explainable metrics.

LLMs, such as GPT-3 \cite{brown2020language} and LLaMA \cite{touvron2023llama}, have shown remarkable summarization capabilities. They can generate coherent and contextually grounded summaries. This makes them ideal for evaluation purposes. LLMs provide both interpretability and inherent summarization abilities. They can generate explanations and reasoning for their predictions, giving evaluators a deeper understanding of system strengths and weaknesses.

Moreover, LLMs reduce the dependency on gold-standard reference summaries. By using LLMs as evaluators, we can generate comparative summaries and objectively assess system-generated summaries. This eliminates potential biases from human references.

In summary, using LLMs as explainable metrics in summarization evaluation offers several benefits. It overcomes the limitations of traditional methods, provides interpretability, and reduces reliance on human-generated references. This emerging field of research holds promise for a more comprehensive and objective assessment of summarization systems. 

The main contribution of this paper is the investigation of various prompt-based methods for explainable evaluation of summarization tasks. We explore both few-shot and zero-shot approaches in our experiments. The best performance prompt follows the zero-shot strategy and is introduced in the paper under the name P1. In this prompt, the criteria for assessing the quality of summaries are described (e.g., how well the main idea of the main document is captured in the summary). This prompt achieves a Kendall correlation score of 0.477, outperforming other methods in comparison.  Our conducted experiments highlight the promising potential of LLMs as evaluation metrics in the field of NLP, with a specific focus on summarization.

\section{Related Work}
Several recent studies have focused on utilizing LLMs for the evaluation of several different tasks in NLP (e.g., text generation, machine translation, and summarization).

GPTScore \cite{fu2023gptscore} is a novel framework for evaluating generated texts using large pre-trained language models, particularly GPT-3. The framework leverages the emergent abilities of generative pre-trained models, such as zero-shot instruction, to score generated texts. GPTScore operates under the assumption that a large pre-trained language model is more likely to assign higher probabilities to high-quality generated text when provided with adequate instruction and context. By leveraging the power of GPT-3, GPTScore aims to assess the quality of generated text based on the model's generative capabilities.

In a similar vein, \citet{wang2023chatgpt} conducted a preliminary survey on using ChatGPT, a variant of the GPT model, as a natural language generation (NLG) evaluator. The study explores the potential of ChatGPT in evaluating the quality of generated text in various NLG tasks.

In the context of translation quality assessment, GEMBA \cite{kocmi-federmann-2023-large} is introduced as a GPT-based metric that can effectively evaluate translations with or without a reference translation. The evaluation focuses on zero-shot prompting and involves comparing four prompt variants in two modes, depending on the availability of a reference. Results from the evaluation demonstrate that GEMBA achieves state-of-the-art accuracy when compared to MQM-based human labels, as evidenced by the WMT22 Metrics shared task.

Instructscore \cite{xu2023instructscore} is an open-source and explainable evaluation metric for text generation. This model fine-tunes the LLaMA model to predict a fine-grained error diagnosis of machine translated content. This work presents a novel framework for explainable text generation evaluation, addressing the limitations associated with black-box metrics and showcasing the potential of LLMs to provide meaningful and interpretable evaluations.

G-Eval \cite{liu2023geval}, is a framework that utilizes LLMs with chain-of-thoughts (CoT) and a form-filling paradigm to assess the quality of NLG outputs, specifically in text summarization and dialogue generation tasks. The experiments demonstrate that G-Eval, utilizing GPT-4 as the backbone model, achieves a high Spearman correlation of 0.514 with human evaluations in the text summarization task, outperforming previous methods significantly.

\section{Task Description}
The topic of  Eval4NLP shared task \cite{eval4nlp23} is to provide explainable metrics for summarization and machine translation evaluation by prompting LLMs. The main goal is to investigate different prompting methods (e.g., zero-shot, few-shot, Chain of Thought, Fine-Grained, Majority Vote, Self-Refinement), 
therefore, \textbf{fine-tuning the LLMs is not allowed}. Also, a number of LLMs are allowed to be used. 
The shared task has two tracks based on the model sizes (One for models bigger than 25B parameters, and one for smaller models). 

This work has been done on the summarization task and using small models. In the following, the dataset and the evaluation metric are explained. 

\subsection{Data}
The Shared Task organizers opted for SummEval during the training and development phase for summarization.  
\citet{10.1162/tacl_a_00373}  introduced SummEval as an evaluation benchmark, aiming to compare various methods for assessing summarization. This benchmark entails the assignment of human ratings on four key dimensions of every summary, including \texttt{fluency, coherence, consistency, and relevance}. SummEval draws upon the renowned CNN/DailyMail dataset proposed by \citet{hermann2015teaching} for its construction.

 Furthermore, in the testing phase, a new reference-free dataset with summary-level quality scores is collected for summarization. 
As source data, sentences and paragraphs were collected from English Wikipedia pages created after 15.07.2023. Test-phase scores are constructed from fine-grained scores.

\subsection{Evaluation Metrics}
To determine how well LLMs explainable metrics correlate with humans, We follow the evaluation protocol of the WMT22 metrics shared task. we use Kendall’s Tau correlation \cite{freitag-etal-2022-results}. In addition to Kendall correlation, Pearson \cite{PMID:23638278} and Spearman \cite{doi:https://doi.org/10.1002/0470011815.b2a15150} are also used in the test phase.

\section{Methodology}

In this research, We have used pre-trained
 orca\_mini\_v3\_7b (tuned Llama2-7b model) \cite{orca_mini_v3_7b} on the HuggingFace Transformers\footnote{\href{https://huggingface.co/pankajmathur/orca_mini_v3_7b}{https://huggingface.co/pankajmathur/orca\_mini\_v3\_7b}}. We employed two strategies, zero-shot and few-shot, for constructing prompts. 
 
\textbf{The zero-shot strategy }included the combination of evaluation criteria for the quality of summarization in the form of questions or detailed explanations provided to the model.

There are examples of summarization evaluation written in \textbf{the few-shot strategy}. In this way, the main document, the summarized document, and the scores received are mentioned precisely. 

To assess summarization quality via prompting an LLM,
the following parameters are needed:
\begin{itemize}
    \setlength\itemsep{-0.4em}
    \item Prompt variant (from a pre-defined set)
    \item Main document \texttt{Source Text}
    \item Summary \texttt{Summary}
\end{itemize}
\subsection{Prompt variants}

For \textbf{P1} (Table~\ref{p1}), we formulated the main criteria for assessing summary quality, which were originally expressed by humans. In this prompt, the following items are mentioned to be checked:
\begin{itemize}
    \setlength\itemsep{-0.4em}
    \item Comparing the key points.
    \item Capturing the main ideas.
    \item Score on a continuous scale from 0 to 100.
    \item Meaning of zero score: irrelevant, factually incorrect, and not readable summary.
    \item Meaning of a hundred score: relevant, factually correct, good readability summary.
    \item Explain the result.
\end{itemize}
To create prompt \textbf{P2}(Table~\ref{p2}), we consulted the ChatGPT4 Bot and asked what questions would be relevant for evaluating summarization. We then modified and adapted the generated questions accordingly. In this prompt, in addition to the items mentioned in P1, the following items have been added in the form of questions.
\begin{itemize}
    \setlength\itemsep{-0.4em}
    \item The overall length of the summary. Concise representation of the original document.
    \item Grammatical accuracy and fluency of the summary.
    \item Evaluate The ranking of information in the summary.
    \item Analyze the level of abstraction in the summary.
    \item Contextual understanding is exhibited by the summary.
\end{itemize}
Prompts P1 and P2 also include an Explanation for the model's output score, thus containing questions that aid in better understanding the received score's reasoning.

Prompt \textbf{P3}(Table~\ref{p3}) and \textbf{P4}(Table~\ref{p4}) are similar to the P1 prompts,  and only the wording and the way of expression have changed.

In \textbf{P5} (Table~\ref{p5}), we guide the model to calculate the desired score by calculating the similarity of the main and summarized documents. P5 includes examples of how one can calculate the similarity of two documents.

Prompt \textbf{P6} (Table~\ref{p6}) follows the few-shot strategy, where two examples consisting of the main document, and the written summary, along with their respective score, are included in the input prompt.

\section{Results and Analysis}

We experiment with six different 
 distinct prompt types. One of them is few-shot (P6) and the rest are zero-shot. Table~\ref{Result} shows results
for all prompt variants we have experimented with.

 \begin{table}[h!]
    \centering
    \begin{tabular}{c | c c c}
        \hline
         & \ Kendall & Pearson & Spearman \\
        \hline
         P1 & \textbf{0.477} & 0.495 & \textbf{0.619} \\
         P2 & 0.470 & 0.468 & 0.607 \\  
         P3  & 0.472 & 0.498 & 0.612 \\
         P4  & 0.467 & 0.504 & 0.610 \\
         P5 & 0.454 & \textbf{0.543} & 0.589 \\
         P6 & 0.283 & 0.513 & 0.376 \\
        
        \hline
        \hline
    \end{tabular}
    
    \centering
    \caption{\label{Result} Test phase Segment-level Kendall ($\tau$) and Pearson ($\rho$) and Spearman correlation.}
    
\end{table}

 \begin{table*}
\centering
\small
\begin{tabular}{p{0.8\linewidth}}
\toprule
\textbf{Prompt  P1: } 
\\
Score the effectiveness of the summarization by \textbf{comparing the key points and overall coherence} of the summarized with the main document.\\\\
Checked whether the summary \textbf{captures the main ideas, maintains the intended tone and style, and provides a concise yet comprehensive
overview} of the main document.
\\\\
Score the summarization \textbf{with respect to the summarized document}
on a \textbf{continuous scale from 0 to 100}, where a score of zero means
\texttt{irrelevant, factually incorrect and good readable} and score of 100 means
\texttt{relevant, factually correct, no readability} summarized.\\\\
Also explain your process to get this score to summary. \\\\
Also please perform error Analysis of given summary.\\\\
What should we change to have a better result?",\\\\
main document: \{main document\},\\\\
Summary: \{Summary\}",\\\\
\textbf{Score}: \textit{gen 'score' pattern='(100|[1-9]?[0-9])'},\\\\
\textbf{Explanation}: \textit{gen 'explanation'}\\\\
        \hline
        \bottomrule
    \end{tabular}

    \caption{The best-performing prompt based on zero-shot prompting strategy expecting a score between 0–100.}
        \label{p1}
\end{table*}

The execution of each prompt takes approximately 1 hour. If we also include the explanation of the results in the output, each execution of the test data takes close to 4 hours.

Based on the Kendall measure (which serves as the primary evaluation metric), the best result is associated with P1. This prompt follows the zero-shot strategy. In this prompt, some of the SummEval criteria are also mentioned. Additionally, P1 achieves the highest value in the Spearman measure and serves as the final strategy for the test phase.

The results of P2, P3, and P4 are very close to each other. The reason for the difference observed is the variation in the way the evaluation method is expressed. In this regard, it can be said that LLMs are sensitive to manner of expression. However, considering the proximity of the Kendall output value, it can be concluded that they have a low sensitivity to the mentioned changes.

Furthermore, considering the results of P5, it can be stated that introducing scientific methods for examining the similarity between summaries and the main document did not effectively guide the model. Instead, criteria such as "captures the main ideas" yielded better results.

Contrary to our expectations, P6 (few-shot approach) obtains the lowest score in the Kendall measure. We expected that the few-shot strategy would outperform zero-shot since it allows the model to observe multiple instances of scoring, thereby enhancing its capabilities. However, our experiments yielded results contrary to this assumption. There may be several reasons for this result. It is possible that a larger number of input samples would have been beneficial. Furthermore, the quality of the input samples might not have been sufficient for the model to comprehend the problem-solving process fully.

In conclusion, based on the obtained results, it can be inferred that by explicitly defining evaluation metrics, language models can be utilized as an interpretable method for evaluating the summarization task.

\section{Conclusion}
In this paper, we have investigated different prompts to define explainable evaluation metrics for summarization Using LLMs.

The experiments conducted indicate that LLMs have great potential as evaluation metrics in NLP tasks, especially summarization. In these experiments, both the few-shot and zero-shot approaches were used. Our best prompt achieved a Kendall correlation of 0.477 compared to the human score. 

In future work, other prompt strategies, such as the Chain of Thought, can also be explored. Experiments can also be repeated with existing prompts and other Language Models and compare the results obtained to determine the effect of the Language Model on changing the quality of the output. 

\section{Limitations}

Due to hardware limitations, we were unable to investigate other eligible models in this series of experiments. In future research, it would be beneficial to examine the impact of other models on the introduced prompts more extensively.
Furthermore, the lack of fine-tuning the model on the defined tasks may have an effect on its performance. In future research, by fine-tuning the model, we can explore its impact on improving the output quality.
\bibliography{anthology,custom}

\appendix
\section{Appendix: Prompt Templates}

Below, we present our prompt templates utilized in the described experiments in this paper.

\begin{table*}[h!]
\centering

\begin{tabular}{p{0.8\linewidth}}
\toprule
\textbf{Prompt  P2: } 
\\
Score the effectiveness of the summarization by \textbf{comparing the key points and overall coherence} of the summarized with the main document.\\\\
Checked whether the summary \textbf{captures the main ideas, maintains the intended tone and style, and provides a concise yet comprehensive
overview} of the main document.
\\\\
Score the summarization \textbf{with respect to the summarized document}
on a \textbf{continuous scale from 0 to 100}, where a score of zero means
\texttt{irrelevant, factually incorrect and  no readability} and score of 100 means
\texttt{relevant, factually correct, good readable} summarized.\\\\
To calculate Score, first answer the following questions.\\
Then, according to the answers to the questions, scored the quality between 0 and 100.\\
1. Assess the overall length of the summary. Does it provide a concise representation of the original document without omitting important information?\\
2. Examine the grammatical accuracy and fluency of the summary. Are the sentences well-structured, free of errors, and coherent?\\
3. Evaluate the ranking of information in the summary. Are the most salient and crucial details given appropriate emphasis and positioned prominently?\\
4. Analyze the level of abstraction in the summary. Does it effectively distill complex ideas and concepts into more accessible and simplified language?\\
5. Consider the contextual understanding exhibited by the summary. Does it demonstrate an understanding of the original text beyond simple keyword extraction?\\
\\
Also explain your process to get this score to summary. \\\\
Also please perform error Analysis of given summary.\\\\
What should we change to have a better result?",\\\\
main document: \{main document\},\\\\
Summary: \{Summary\}",\\\\
\textbf{Score}: \textit{gen 'score' pattern='(100|[1-9]?[0-9])'},\\\\
\textbf{Explanation}: \textit{gen 'explanation'}\\\\
        \hline
        \bottomrule
    \end{tabular}
    \caption{Prompt P2}
    \label{p2}
\end{table*}

\begin{table*}[h!]
\centering
\begin{tabular}{p{0.8\linewidth}}
\toprule
\textbf{Prompt  P3: } 
\\
Your Task is to score the Samaritan quality. The original document is collected from English Wikipedia pages created after 15.07.2023.\\\\
Score the effectiveness of the summarization by comparing the key points and overall coherence of the summarized with the main document.\\\\
"Checked whether the summary captures the main ideas, maintains the intended tone and style, and provides a concise yet comprehensive overview of the main document.\\\\
Score the summarization \textbf{with respect to the summarized document}
on a \textbf{continuous scale from 0 to 100}, where a score of zero means
\texttt{irrelevant, factually incorrect and not readable} and score of 100 means
\texttt{relevant, factually correct, good readability, grammatical correctness,  covers the main topic and key points of the main document article} \\\\
Source text: \{main document\},\\\\
Summary: \{Summary\}",\\\\
\textbf{Score}: \textit{gen 'score' pattern='(100|[1-9]?[0-9])'},\\\\
        \hline
        \bottomrule
    \end{tabular}
    \caption{Prompt P3}
    \label{p3}
\end{table*}

\begin{table*}[h!]
\centering
\begin{tabular}{p{0.8\linewidth}}
\toprule
\textbf{Prompt  P4: } 
\\
Score the effectiveness of the summarization by comparing 
 the key points and overall coherence of the summarized with 
 the main document.\\\\
Checked whether the summary captures the main ideas, 
maintains the intended tone and style, 
and provides a concise yet comprehensive overview of the main document.\\\\
Score the summarization with respect to the summarized document,
on a continuous scale from 0 to 100.\\\\
Source text: \{main document\},\\\\
Summary: \{Summary\}",\\\\
\textbf{Score}: \textit{gen 'score' pattern='(100|[1-9]?[0-9])'},\\\\
        \hline
        \bottomrule
    \end{tabular}
    \caption{Prompt P4}
    \label{p4}
\end{table*}

\begin{table*}
\centering
\begin{tabular}{p{0.8\linewidth}}
\toprule
\textbf{Prompt  P5: }\\
Score the summarization with respect to the summarized document on a \textbf{continuous scale from 0 to 100}, where a score of zero means
\texttt{irrelevant, factually incorrect and not readable} and score of 100 means \texttt{relevant, factually correct, good readability}.\\
let's think step by step.\\
In other words, this Score should show the similarity between the main document and the summarized document.\\
For similarity measurement, It's possible to compare the main and summarized document with a similarity measure such as Cosine Similarity.\\
word2vec, Bert embedding or n-gram are some of the approaches to calculate similarity.\\
                
Source text: \{main document\},\\\\
Summary: \{Summary\}",\\\\
\textbf{Score}: \textit{gen 'score' pattern='(100|[1-9]?[0-9])'},\\\\
        \hline
        \bottomrule
    \end{tabular}
    \caption{Prompt P5}
    \label{p5}
\end{table*}
\begin{table*}[h!]
\centering
\small
\begin{tabular}{p{0.8\linewidth}}
\toprule
\textbf{Prompt  P6: } 
\\
\textbf{Consider these example that summarization is graded in scale 0 - 100.}\\\\
1. \textbf{Source text:} Usain Bolt will compete at the IAAF/BTC World Relays in the Bahamas next month, the Jamaica Athletics Administrative Association has announced. The six-time Olympic gold medallist will compete at the relay championship on May 2 and 3 as part of the Jamaican team. 'I'm happy to be part of the Jamaican team for the IAAF / BTC World Relays in the Bahamas. I am fit, healthy and ready to run,' said Bolt. Usain Bolt has confirmed he will be part of Jamaica's team at the World Relays in the Bahamas Bolt reacts as he wins 4x100m gold at the London Olympic Games in 2012 'I hear the meet was a lot of fun last year and there was a great atmosphere. Jamaica has a long and successful tradition in relays and when we put on the national colours we always do our best to make the country proud,' he added. JAAA General Secretary Garth Gayle commented, 'We were extremely pleased that Usain was available for selection and that the world's fastest man will be running for Jamaica. We can expect some sprint magic on the track in the Bahamas on 2nd and 3rd May.' The full Jamaican team list for the competition will be announced shortly. Bolt insists he always does 'his best to make his country proud' while wearing Jamaica colours.\\\\
1. \textbf{Summary:} Jamaican sprinter Usain Bolt has confirmed he will be part of the Jamaican team at the IAAF/BTC World Relays in the Bahamas.\\
1.\textbf{ Score : 95,}\\\\
2. \textbf{Source text:} Referee Mark Clattenburg has been named to take charge of the Manchester derby on Sunday, despite having sent off three players from United and City this season. City captain Vincent Kompany was dismissed for two bookable offences during Belgium's narrow 1-0 defeat of Israel in their Euro 2016 qualifier on March 31, meaning he is now suspended for the match against Wales in June. And, although Clattenburg has been accused of favouring Louis van Gaal's side in the past, it's worth noting that the 40-year-old has only sent off two players season in the Premier League this season and both have been from United; Tyler Blackcett in the 5-3 defeat by Leicester and Luke Shaw in the 1-1 draw with West Ham. Mark Clattenburg will officiate the Manchester derby between United and City at Old Trafford The English referee sent off City and Belgium captain Vincent Kompany during the international break Leicester 5-3 Manchester United West Ham 1-1 Manchester United Manchester United 3-0 Tottenham Manchester City 3-1 West Ham Liverpool 2-1 Manchester City Chelsea 1-1 Manchester City Clattenburg has courted controversy during his career but is generally regarded as one of the Premier League's leading referees alongside Michael Oliver. The champion's shock 2-1 loss to Crystal Palace on Monday saw United move a point above their local rivals to add extra incentive for both sides ahead of the derby at Old Trafford, which could ultimately decide who finishes second behind expected winners Chelsea. While Manuel Pellegrini's side have struggled since the turn of the year, turning from title challengers to fourth place chases, United are coasting on confidence having won their last five consecutive league games. Clattenburg will be joined on Sunday by assistants Simon Beck and Jake Collin, while Jonathan Moss will serve as the fourth official. Clattenburg has shown only two red cards this season, both to United players including Luke Shaw (centre).\\\\
2. \textbf{Summary: }United's win over Liverpool was their first league win since the 3-0 win over Leicester on March 31 City's win over West Ham was their first league win since the 3-0 win over Chelsea on March 31 Manchester City's win over West Ham was their first league win since the 3-0 win over Chelsea on March 31 Manuel Pellegrini's side are top of the Premier League table, four points clear of Chelsea, who have a game.\\
2. \textbf{Score : 26.666666666}\\\\
\textbf{following these examples, please score the following input}.\\\\
Source text: \{main document\},\\\\
Summary: \{Summary\}",\\\\
\textbf{Score}: \textit{gen 'score' pattern='(100|[1-9]?[0-9])'},\\\\
        \hline
        \bottomrule
    \end{tabular}
    \caption{Prompt P6}
    \label{p6}
\end{table*}

\end{document}